\documentclass{article}

\usepackage{array}
\usepackage{longtable}
\usepackage{arxiv}
\usepackage{hanging}
\usepackage[utf8]{inputenc} 
\usepackage[T1]{fontenc}    
\usepackage{hyperref}       
\usepackage{url}            
\usepackage{booktabs}       
\usepackage{amsfonts}       
\usepackage{nicefrac}       
\usepackage{microtype}      
\usepackage{lipsum}		
\usepackage{graphicx}
\usepackage{natbib}
\usepackage{doi}
\usepackage{longtable}
\usepackage{multirow}
\usepackage{lscape} 
\usepackage{academicons}
\usepackage{xcolor}
\usepackage{authblk}

\newcommand{\orcid}[1]{\href{https://orcid.org/#1}{\textcolor[HTML]{A6CE39}{\aiOrcid}}}

\author[1,2]{Xiaoming Zhai \thanks{Corresponding author.
Address: 125M Aderhold Hall, 110 Carlton Street 
Athens, Georgia 30602
Office: 517-432-0816
Email: Xiaoming.Zhai@uga.edu}  }

\author[1,2]{Matthew Nyaaba}
\author[3]{Wenchao Ma}

\affil[1] {AI4STEM Education Center, University of Georgia,Athens, GA, USA }
\affil[2] {Department of Mathematics, Science, and Social Studies Education, University of Georgia,Athens, GA, USA}
\affil[3]{College of Education, University of Alabama, Tuscaloosa, AL, USA }

\title{Can Generative AI and ChatGPT Outperform Humans on Cognitive-demanding Problem-Solving Tasks in Science?}

\renewcommand{\headeright}{Zhai, Nyaaba \& Ma}
\renewcommand{\undertitle}{Paper Accepted to Science \& Education}
\renewcommand{\shorttitle}{GPT Outperforms Human on Problem Solving}

\hypersetup{
pdftitle={A template for the arxiv style},
pdfsubject={q-bio.NC, q-bio.QM},
pdfauthor={David S.~Hippocampus, Elias D.~Striatum},
pdfkeywords={First keyword, Second keyword, More},
}

\begin{document}
\maketitle

\begin{abstract}
This study aimed to examine an assumption regarding whether generative artificial intelligence (GAI) tools can overcome the cognitive intensity that humans suffer when solving problems. We examine the performance of ChatGPT and GPT-4 on NAEP science assessments and compare their performance to students by cognitive demands of the items. Fifty-four 2019 NAEP science assessment tasks were coded by content experts using a two-dimensional cognitive load framework, including task cognitive complexity and dimensionality. ChatGPT and GPT-4 answered the questions individually and were scored using the scoring keys provided by NAEP. The analysis of the available data for this study was based on the average student ability scores for students who answered each item correctly and the percentage of students who responded to individual items. The results showed that both ChatGPT and GPT-4 consistently outperformed most students who answered each individual item in the NAEP science assessments. As the cognitive demand for NAEP science assessments increases, statistically higher average student ability scores are required to correctly address the questions. This pattern was observed for Grades 4, 8, and 12 students respectively. However, ChatGPT and GPT-4 were not statistically sensitive to the increase of cognitive demands of the tasks, except for Grade 4. As the first study focusing on comparing cutting-edge GAI and K-12 students in problem-solving in science, this finding implies the need for changes to educational objectives to prepare students with competence to work with GAI tools such as ChatGPT and GPT-4 in the future. Education ought to emphasize the cultivation of advanced cognitive skills rather than depending solely on tasks that demand cognitive intensity. This approach would foster critical thinking, analytical skills, and the application of knowledge in novel contexts among students. Furthermore, the findings suggest that researchers should innovate assessment practices by moving away from cognitive intensity tasks toward creativity and analytical skills to more efficiently avoid the negative effects of GAI on testing. 

\end{abstract}


\newpage
\section{Introduction}

Humans face increasing challenges in solving problems or designing solutions if the tasks are more cognitively demanding. This is largely due to the limited working memory capacity for processing complex problems and serves as a fact that has limited humans’ ability to solve science problems and learn science (Paas et al., 2003). With the development of Generative Artificial Intelligence (GAI) such as ChatGPT, it is  that GAI can solve problems that humans were not able to (Latif et al., 2023), especially for those that were time-pressing and cognitively demanding. It is thus critical to examine whether cutting-edge GAI tools outperform humans in problem-solving, particularly in overcoming cognitively demanding tasks. Addressing this question is imperative in the current landscape where GAI is ubiquitous in education and every sector of society, because the answer, better or worse, may alter the educational goals set up to enhance future financial and societal advances in the rapidly changing world. 

In traditional educational settings, improving students’ knowledge and ability to solve increasingly cognitively demanding problems has always been one of the critical educational goals. High cognitive demanding tasks usually require students to make connections, analyze and evaluate evidence, and draw conclusions, and are considered essential for students to develop a deeper understanding of science (Estrella et al., 2020). Students educated to possess such fundamental skills would be more likely to succeed in their future careers, where precedent challenges may occur, and this essential skill is needed to meet the challenge. Therefore, educational systems have been consciously designed to help students achieve this goal. For example, large cross-national studies found that countries that perform well on international science tests, like Singapore, have textbooks designed specifically to improve a variety of cognitive skills for students rather than simply emphasizing recall and comprehension (Johnson \& Boon, 2023). Such textbooks and others provide students with various opportunities to improve their cognitive skills by using scientific knowledge to solve problems. Included in these learning materials are some higher-cognitive-demanding tasks, which can be challenging for students to perform and are supposed to prepare students with the essential competence for career success in the future. Many educators contend that cognitively intensive tasks are essential to improve student's learning; others argue that cognitively intensive tasks pose additional pressure to students without known benefits. 

With the surge of GAI, researchers and educators need to reconsider this problem. Does education need to rely on high cognitive intensity tasks to promote efficient learning if GAI can overcome such barriers? Prior research has shown the powerful problem-solving ability of ChatGPT. For example, Zeng (2023) examined the problem-solving ability of ChatGPT and found that it was accurate on tasks requiring common sense and knowledge. Orrù et al. (2023a) compared ChatGPT with 20 humans in solving both practice problems and transfer problems and concluded that ChatGPT was in line with the mean performance of humans. These results show some potential for GAI to overcome cognitively demanding issues when solving problems.  

Should GAI become capable of surpassing human performance in tasks that require significant cognitive effort, it could necessitate an immediate and crucial dialogue regarding the reconsideration of our educational objectives. In essence, the domain must introspectively consider and pinpoint the fundamental abilities that will be crucial for the workforce of the future, one that will function in an GAI-dominated environment. These skills may extend beyond traditional knowledge and technical expertise, encompassing abilities that will allow individuals to coexist and collaborate effectively with GAI systems (Zhai, 2022), such as critical thinking, creativity, emotional intelligence, and adaptability, among others. We must prioritize these skills, positioning them at the forefront of our educational goals. By doing so, we can ensure that future generations are well-prepared and adept at navigating the ever-evolving technological landscape. This not only prepares them for the working world but also empowers them to contribute effectively to society at large. Therefore, the advent of GAI and its impacts on education should be closely examined, and based on the outcomes, educators must redefine and refine the education goals and the approach to education, ensuring they remain relevant and effective in the age of GAI (Zhai, 2023). 

This study represents one such effort. We aim to examine the most powerful GAI tools’ (i.e., ChatGPT and GPT-4) performance on cognitively demanding tasks with the US National Assessment of Educational Progress (NAEP) science items. NAEP science is rigorously designed to provide a fair and accurate measure of US students’ academic achievement that reflects trends (Zhai \& Pellegrino, 2023). The tasks are designed with varying levels of cognitive demands that require K-12 students to use scientific knowledge to identify science principles, use science principles, use scientific inquiry, and use technological design. These assessment tasks thus provide a perfect measure to examine GAI’s ability to solve problems, as well as compare GAI’s ability with that of students. We specifically asked the following research questions:  
\begin{itemize}
    \item a) Can ChatGPT or GPT-4 outperform age-appropriate average humans on NAEP science assessments? 
    \item b) How do ChatGPT and GPT-4 perform on NAEP science assessments by cognitive demand, compared to age-appropriate average students?
\end{itemize}

\section{Cognitive Load on Science Problem Solving}

Cognitive load in science education discusses the mental effort required by students to process and comprehend scientific knowledge and concepts (Sweller's, 2011; Paas, et al., 2003). Sweller's (2011) cognitive load theory was adopted to examine NAEP science assessments, highlighting the different ways GAI and humans handle the cognitive load. Sweller's cognitive load includes three types: Intrinsic Cognitive Load (ICL), Extraneous Cognitive Load (ECL), and Germane Cognitive Load (GCL).  

ICL refers to the inherent difficulty associated with a specific task. In science problem-solving, ICL deals with the complexity of scientific knowledge itself (Gupta \& Zheng, 2020). The NAEP Science Assessments cover a wide range of scientific topics, each with its inherent complexity (Bergner \& von Davier, 2018; Lim \& Sireci, 2017). The ICL of these assessments is tied to the nature of the scientific knowledge being tested, from basic principles in early education to more complex theories in higher grades (Gerjets et al., 2009). This intrinsic load is unchangeable as it is directly linked to the content's complexity. For example, questions on ecosystems or the laws of motion inherently may require a certain level of cognitive processing due to their complexity. 

The ECL is imposed by the way information is presented to learners and can be controlled by task design. This is where the design and structure of the NAEP assessments play a crucial role (Hadie \& Yusoff, 2016). If the assessments are formulated with complex wording, ambiguous instructions, or include unnecessary information, they increase the ECL (Tugtekin \& Odabasi, 2022). This can detract from students’ ability to concentrate on the primary scientific knowledge being assessed. Therefore, ensuring that the assessments are straightforward, clear, and directly focused on the intended scientific principles is key to minimizing ECL and may have an effect on the performance of students (Behmke \& Atwood, 2013).

GCL is concerned with the mental work done by learners to understand and assimilate new information into their existing knowledge base (Paas \& Van Merriënboer, 2020). This type of load is essential for deep learning and is associated with the processing and construction of schemas (Pengelley et al., 2023). In the context of the NAEP Science Assessments, GCL will involve creating questions that not only assess students’ recall of scientific facts but also their ability to apply concepts, analyze data, and think critically (Lagalante, 2023). This could involve scenario-based questions that require students to apply their understanding to novel situations, thereby promoting deeper cognitive processing and understanding. 

Understanding the working memory's role in these assessments is also vital. Working memory's limited capacity means that if an assessment is too high in either intrinsic, extraneous or germane load, it could overwhelm students, leading to poor performance (Feldon et al., 2019). Balancing these loads is crucial in accurately assessing students' true understanding of science (Wang et al., 2017). Working memory is the mental space where information is temporarily held and manipulated. It has a limited capacity, which is why cognitive load theory is so important in designing science assessments. The intrinsic, extraneous, and germane loads all compete for these limited working memory resources (Paas \& Van Merriënboer, 2020).

Numerous scholarly works have investigated the impact of cognitive load on science learning, highlighting the importance of managing cognitive load to optimize learning outcomes and enhance problem-solving skills. It is posited that students’ cognitive processing ability to comprehend science knowledge is limited, and therefore, the teaching and assessment of science require the management of cognitive load demand based on students’ levels and needs (Sweller, 2011). In the context of this study, it is assumed that the concept of working memory is not directly applicable to GAI tools like ChatGPT and GPT-4 as it is to humans.  

Cognitive demand in science focuses on identifying the level and kind of reasoning required of students to successfully engage with a task that focuses on using science knowledge when engaging in scientific practices (Park, 2011). Cognitive demand levels in science range from tasks that require recall (memory) and are in direct correspondence with definitions to using knowledge to conduct tasks (Park, 2011). By categorizing science tasks according to two key dimensions— the integration of science content (i.e., disciplinary core ideas and crosscutting concepts) and practices and the cognitive complexity, Tekkumru‐Kisa et al. (2015) identified nine distinct categories of cognitive activities. The first dimension refers to integration (that is, scientific content, scientific practices, or the combination of the two). The second dimension, cognitive complexity, includes five levels. In the lowest two levels, memorization and scripted tasks are given respectively. There are few chances for students to engage in critical thinking and reasoning about science content and/or practices in these tasks. In the upper three rows, high-level tasks are given--Levels 3 and 4 contain tasks that require guidance for comprehension, while level 5 tasks involve doing science. 

The concept of cognitive load in science is instrumental to teachers in evaluating and predicting the performance of their students. It demonstrates the varied levels of comprehension among the students. The information that is obtained about the mental processes of students using cognitive load has the potential to shed light on the issue of the variation in learning outcomes that exist across different learners even though they receive the same instruction (Hadie \& Yusoff, 2016). It has been found that many teachers face difficulties in using higher-level cognitively demanding tasks in science assessment, with the level of cognitive demand tending to decline after implementation (McCormick, 2016).




\section{GAI, ChatGPT, and Problem-solving }

ohn McCarthy first proposed AI in 1956 at the Dortmund conference. AI appears to have received much attention recently due to the development of machine learning (Jordan \& Mitchell, 2015). A variety of definitions exist for AI due to its polysemy nature (Zhai et al., 2020). Nonetheless, many definitions center on AI as a computer-based tool that simulates human-like thinking, including reasoning, argumentation, learning, sense-making, communication, analysis, decision-making, and generalization. This describes how AI mimics human intelligence. The most recent agent of AI that has received much attention in education is the GAI chatbots (i.e., ChatGPT and GPT-4) developed by OpenAI (2022). Chatbots are considered one of the most advanced AI systems that can solve difficult problems with greater accuracy due to their broader general knowledge and problem-solving abilities. 

ChatGPT inherits its architecture from the GPT-3.5 model and leverages a transformer-based architecture consisting of numerous layers of self-attention and feed-forward neural networks (Assaraf, 2022). The model is trained on a massive corpus of text data in two steps: pretraining and fine-tuning. The pretraining process used a large amount of publicly available text data, allowing the model to learn a wide range of syntactic and semantic features. Fine-tuning is performed on specific conversational datasets, enabling the ChatGPT to specialize in a dialogue context. To optimize the model, reinforcement learning was employed based on human feedback. GPT-4 is an upgraded version of ChatGPT with more reliability and creativity to handle more nuanced tasks (OpenAI, 2023).  

The ability of ChatGPT to respond to questions or generate responses is a form of problem-solving that has the potential to benefit various fields, including education (Zhai, 2023). Research on GAI and problem-solving explores various aspects of GAI, its applications, challenges, and impact on different areas of life, with a focus on how GAI can be used to solve complex problems by imitating human behavior using advanced algorithms and techniques (Kung et al., 2023). However, the quality of these responses is crucial in determining the effectiveness of GAI chatbots in problem-solving. For examples, the study conducted by Li et al. (2023) tested ChatGPT's accuracy in answering questions from Taiwan's educational examinations. ChatGPT demonstrated a high accuracy rate, often exceeding 80\%, indicating its effectiveness as a learning tool in mathematics education. This suggests that ChatGPT can significantly aid in improving self-regulation and potentially revolutionize middle school math education. On the other hand, an assessment of ChatGPT versions PT-3.5 and GPT-4 in solving biostatistical problems indicated a below-average performance on initial attempts, though GPT-4 managed to provide all correct answers within three attempts with precise guidance.




Furthermore, study by Ignjatović and Stevanović (2023) and Orru et al (2023) focused on ChatGPT's cognitive abilities, particularly in problem-solving and verbal insight tasks (Orrù et al., 2023b). The findings revealed that ChatGPT could match human performance in practice and transfer problems, aligning with the most probable outcomes expected from a human sample. This parity in performance underscores ChatGPT's capability to mirror the average success rate of human subjects, thereby showcasing its proficiency in cognitive tasks. 

Williams (2023) and Seetharaman (2023) also presented insights into the broader implications of ChatGPT's capabilities. Williams for instance, emphasized the challenges in detecting and aligning advanced artificial general intelligence (AGI) with collective well-being, while Seetharaman highlighted ChatGPT's potential in education, enhancing skills essential for practice and critical thinking. These varied studies collectively underscore ChatGPT's adeptness in cognitive functions and its promising role in diverse educational and practical applications. 

In light of the prior findings, it is important to consider the level of thinking that chatbots can produce, including higher-order thinking, common thinking, and creative solutions (Bang et al., 2023). While chatbots have demonstrated the ability to generate responses that are comparable to those of humans in some cases, the extent to which they can produce creative and original solutions to science problems remains an open question.  

\subsection{Higher-order Thinking }
GAI technologies have revolutionized problem-solving by enabling the analysis of vast amounts of data with high levels of certainty (Daher et al., 2023). Deep learning algorithms are often utilized to search for patterns in large datasets, enabling GAI to detect complex relationships that may be difficult for humans to uncover (Bergen et al., 2019; Najafabadi et al., 2015). ChatGPT has become increasingly important recently as a means of applying GAI to solving higher-order thinking problems. For instance, the study conducted by Sinha et al. (2023) provides important insights into the potential of GAI in solving higher-order problems in pathology. This is an indication that this level of cognition in GAI has the potential to be a useful tool for students and academics alike, enabling them to develop critical thinking abilities if effectively utilized (Latif et al., 2023).  

\subsection{GAI and Common-sense Problem-solving }
Common sense refers to the knowledge that is generally accepted as true and accessible to everyone (Rosenfeld, 2011). Common sense includes basic knowledge about the natural world, social norms, and the ability to make judgments and decisions based on experience. The study by Bian et al. (2023) on ChatGPT’s ability to solve commonsense problems highlights that while GAI can achieve good accuracy in commonsense questions, it still struggles with certain types of knowledge, including social and temporal commonsense. The study suggests that ChatGPT is an inexperienced commonsense problem solver and often generates knowledge with a high noise rating. While it is true that open GAI-powered chatbots have the potential to solve a wide range of problems with multimodalities (Cao et al., 2023), researchers have also acknowledged that many challenges still exist that require further study and development (Aktay et al., 2023; Bian et al., 2023). Further research and development are necessary to address these issues and realize the full potential of OpenAI chatbots in solving real-world problems in the education (Adiguzel et al., 2023).

\section{Methods }
\subsection{Instruments and Materials }

In order to examine the ability of ChatGPT and GPT-4 to solve science problems as compared to students, we adopted 54 released 2019 NAEP science assessments, including test items for grade levels 4, 8, and 12. The NAEP assessments are conducted periodically in reading, mathematics, science, writing, U.S. history, civics, geography, and other subjects (Zhai \& Pellegrino, 2023). We chose the science test because science uses empirical standards, logical arguments, and skeptical review to distinguish itself from other ways of knowing and thinking (NGSS Lead States, 2013). In the context of the NAEP science assessment, the subject domain content is integrated with four types of science practices to form performance expectations. The three domain content areas are physical science, life science, and earth and space sciences (National Assessment Governing Board, 2019), while the science practices include identifying science principles, using science principles, using scientific inquiry, and using technological design. NAEP items assess students’ use of knowledge when involved in these science practices. Moreover, it is believed that the test items defined the cognitive load experienced by students, and therefore a well-designed assessment would aim to allow students to focus on demonstrating their knowledge and skills in the assessed areas (Brüggemann et al., 2023; Prisacari \& Danielson, 2017). These features made NAEP items the most appropriate for us to test our hypothesis about GAI. 

The 2019 NAEP test is the latest and most openly available data as of the time of this study. Unlike previous NAEP science items, the 2019 NAEP contained a key recommendation on the distribution of questions by science content area and grade, with equal weight given to all three science disciplines in grade 4, an emphasis on Earth and Space Sciences in grade 8, and an emphasis on Physical Science and Life Science at grade 12. This recommendation provided a comprehensive view of science examinations across all K-12 grade levels and enabled the researchers to fully evaluate the effectiveness of ChatGPT and GPT-4 in problem-solving. The 2019 assessment saw the participation of representative samples of students from across the country, including 30,400 4th graders from 1,090 schools, 31,400 8th graders from 1,070 schools, and 26,400 12th graders from 1,760 schools. 

Using the NAEP Question Tools (The Nation’s Report Card, 2022), we collected all the available assessment items for each grade and for all content classifications; a sum total of 33 multiple-choice questions, a sum total of 4 selected response items, a total of 3 scenario-based task questions, a sum total of 11 short constructed response questions, and a sum total of 3 extended constructed response questions, making a grand total of 54 items(an overview of the assessment items refers to Table 1) Multiple choice serves as the principal format in assessments. This paradigm poses a question, or a statement accompanied by a finite range of potential responses, typically extending between three and five options, with only one being accurate. The assessment objective revolves around evaluating the student’s recollection and comprehension of specific subject matter. In comparison to multiple choice, selected responses resemble multiple choice in terms of providing an array of potential answers. However, the distinguishing factor lies in the fact that these may accommodate multiple correct responses. The utility of such items is to appraise the student’s capacity for analysis, evaluation, and application of knowledge as opposed to simple recollection. A scenario-based task is a unique category of assessment that necessitates the application of acquired knowledge and skills to a practical or theoretical circumstance. The student may be presented with a scenario and solicited to resolve a problem, arrive at a decision, or illustrate a process. These tasks often demand interdisciplinary knowledge, thereby assessing higher-order cognitive faculties, namely application, analysis, synthesis, and evaluation. Short-constructed response items demand a written answer, albeit generally confined to a few sentences. The response requires more than mere recollection; students are expected to elucidate, interpret, or apply their knowledge. These items serve to gauge a student’s understanding of a concept, complemented by their capacity for written expression. Extended Constructed Response, as the name suggests, requires a more comprehensive response relative to Short Constructed Response. The response can span from a single paragraph to multiple ones. The evaluative focus is not only on the student’s ability to explain, interpret, or apply knowledge but also on their proficiency in in-depth exploration of a subject matter. This format invariably necessitates higher-order cognitive abilities and offers a more robust platform to assess a student’s writing skills.

\begin{table}[!ht]
    \centering
    \caption{Table 1 Assessment format and number of NAEP 2019 Science }
    \begin{tabular}
    {p{1cm} p{2cm} p{2cm} p{2cm} p{1.5cm} p{1.5cm} p{2cm} p{1cm}}
    \hline
         & Grade level &Multiple Choice (\#) & Selected Responses (\#) &Scenario-Based Task (3) &Short Constructed Response (\#) & Extended Constructed Response (\#)  & Sum (\#) \\ \hline
        Physical Science& 4  & 3 & - & 0 & 3 & 0 & 6 \\ 
        & 8& 3 & 3 & 0 & 0 & 0 & 6 \\ [1ex]
         & 12 & 2 & - & 1 & 2 & 1 & 6 \\
         [1ex]
        total & & 8 & 3 & 1 & 5 & 1 & 18 \\ \hline
        life Science & 4 & 6 & - & 1& 0 & 0 & 7 \\ 
        & 8& 5 & 0 & 0 & 1 & 0 & 6 \\ 
        [1ex]
        & 12 & 6 & - & 0 & 2 & 0 & 8 \\
        [1ex]
        total & & 17 & 0 & 1& 3 & 0 & 21 \\ \hline
        Earth Sciences & 4 & 1 & - & 0 & 1 & 1& 3 \\ 
        &8 & 3 & 1 & 1 & 1 & 1 & 7 \\
        [1ex]
        & 12 & 4 & - & 0 & 1 & 0 & 5 \\ 
        [1ex]
        total &  & 8& 1 & 1 & 3 & 2 & 15 \\ \hline
        Grand &  &  &  &  &  &  &  \\
        Total & sum & 33& 4 & 3 & 11 & 3 & 54 \\ \hline
       
    \end{tabular}
\end{table}

\newpage

\subsection{Coding of the Cognitive Load for the 2019 NAEP Tasks}
\emph{\textbf{Coding Framework and Procedures. } } To address the research questions, three content experts were employed to code the cognitive demand of the 2019 NAEP assessment tasks using a two-dimensional cognitive load framework (NGSS Lead States, 2019), illustrated in Table 2. This framework first categorizes tasks based on their independent cognitive complexity, which reflects the cognitive demand of students’ long-term memory schemes (Tekkumru‐Kisa et al., 2015). This dimension escalates from “Scripted” (Task Complexity 2, TK2) to “Guided” (Task Complexity 3, TK3), and finally to “Doing Science” tasks (Task Complexity 4, TK4). The complexity also augments from integrating two dimensions (e.g., TK2D2, TK3D2, TK4D2) to integrating three dimensions (e.g., TK2D3, TK3D3, TK4D3). The cognitive load of an item is, however, a synthesis of the degree of independence demanded from the student in formulating a response and the extent of dimensional integration.

Items are designed to span this range of cognitive complexities, facilitating a diverse representation of cognitive complexity within each assessment. Notably, an item’s cognitive load comprises two primary components: the level of independence demanded from the student in responding to the item and the degree of dimensionality integration.

For instance, consider this NAEP 2019 Grade Four (4) Science question (Question ID:2019-4S8 \#14 K0783E1) in Figure 1, under the Physical Science (PS) content category with a difficulty level of “Hard,” which required students to determine how sounds and pitch are produced. Specifically, this question presents a science task where a student plucks a rubber band stretched between two nails to make a sound and is asked to explain what makes the sound and how to alter the pitch. Using the framework provided in Table 2, we determined the cognitive load of this task by assessing the dimensions of content engagement required. Following this, the task asked students to apply knowledge of sound generation and modification, which involves understanding the concepts of vibration and tension (how plucking the rubber band generates sound and how increasing tension affects pitch). This required students engaging with two dimensions of content: recognizing how the scientific concept is developed (how sound is produced) and applying this understanding to a practical scenario (altering pitch).

\begin{figure}[htp]
\centering
\includegraphics[width=1\textwidth]{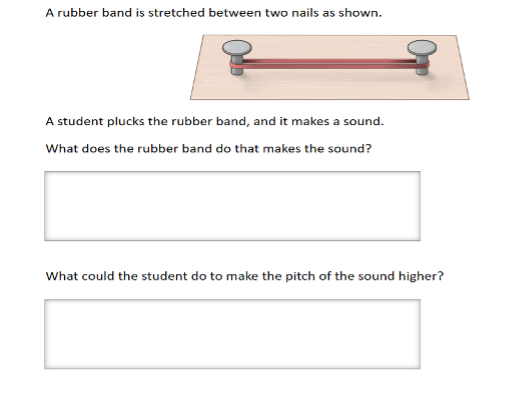}
\caption{Explain how to produce sounds (NAEP, Science, Grade: 4, Year: 2019)}

\end{figure}

Given this description, this task fell into the "Two Dimensions" category of the framework. Within this category, there are three levels of guidance: Scripted Tasks (TK-2D2) where the student is expected to follow a script to work with (or is told how to use) two dimensions to complete a task; Guided Tasks (TK-3D2) provides guidance or scaffolding for students to use two dimensions to complete a task; and Doing Science (TK-4D2), where student engages in two dimensions of content to make sense of and/or recognize how the scientific body is developed. In our coding process, this question does not seem to provide a script or definition for the students to follow, nor does it appear to offer guidance or scaffolding. Instead, students were expected to apply their understanding directly to a scientific phenomenon that involves doing experiments. Therefore, the most appropriate code for the cognitive load of this task seems to be TK-4D2. The student is engaged in doing science by applying two dimensions of content to understand and explain the phenomenon of sound production and modification.


\begin{table}[!ht]

    \caption{A Framework of Cognitive Load for Science Assessment}
      \centering
    \begin{tabular}{p{3cm}p{3cm}p{3cm}p{3cm}}
    \hline
        \multicolumn{2}{c}{Increasing Cognitive Load} \\ [1ex]\hline
        Task/DI & One Dimension & Two Dimensions & Three Dimension\\ [1ex]\hline
        Doing Science 
        
        (TK=4)& N/A-a student cannot “do” science with only one dimension.  & Student engages in two dimensions to make sense of content and/or recognize how the scientific body is developed. 
        
        CODE: TK4D2
 & Student engages in three dimensions to make sense of content and/or recognize how the scientific body of knowledge is developed.
 
 CODE: TK4D3
 \\ \hline
        Guided Tasks

(TK=3) & Student is given some guidance or scaffolding with only a practice to complete OR is provided guidance toward supplying appropriate content as an answer& Student is given some guidance or scaffolding to use two dimensions to complete a task. 

CODE: TK3D2 & Student is given some guidance or scaffolding to use three dimensions to complete a task. 

CODE: TK3D3
 \\ \hline
        Scripted Tasks

(TK=2)& Student follows a script (outline) of a practice OR is told how to use content to solve a problem& Student follows a script to work with (or is told how to use) two dimensions to complete a task.

CODE: TK2D2& Student follows a script to work with (or is told how to use) there dimensions to complete a task. 

CODE: TK2D3
 \\ \hline
        Memorized Tasks

(TK=1)& Student repeats or must provide definition of practices or content & N/A-memorization cannot be completed where integration of dimensions is required & N/A-memorization cannot be complete where integration of dimensions is required.  \\ \hline

    \end{tabular}
\end{table}

The panel of three raters who undertook the coding task possesses expertise in science education in general, with disciplinary expertise in physics, chemistry, and biology, respectively. To ensure uniformity in data interpretation and analysis, all raters were provided with clear guidelines and underwent appropriate training. A few sample items were first selected and assigned to all raters for independent grading. They then discussed the discrepancies till a consensus was met. In this way, they got familiar with the coding framework and ensured their understanding of the coding framework was on the same page. The items were then distributed among the raters for independent coding. The raters used half a day to code the items using the cognitive load framework. To examine the interrater reliability, we calculated the intraclass correlation coefficient (ICC = .978), indicating very high agreement among the raters regarding the cognitive coding of the items (see Table 3).

\begin{table}[!ht]
    
    \caption{Intraclass Correlation Coefficient of The Cognitive Load Rating of the Items}
    \centering
    \begin{tabular}{p{3cm}p{2cm}p{2cm}p{2cm}p{2cm}p{1cm}p{1cm}}
    \hline
         & Intraclass $Correlation^b$ &\multicolumn{2}{c}{95\% Confidence Interval}
         &\multicolumn{3}{c}{F Test with True Value 0} \\ 
         \cmidrule{3-4} \cmidrule{5-7}
         ~ & ~ & Lower Bound & Upper Bound & Value& $df_1$ & $df_2$ \\ \hline
        Single Measures & $.936^a$& .761 & .990 & 44.800 & 5 & 10 \\ 
        [2ex]
        Average Measures& $.978^c$ & .905 & .997 & 44.800 & 5 & 10 \\ \hline
    
    \end{tabular}
\end{table}

 \emph{\textbf{Coding Results} } . Table 4 shows the coding results for each item with the two dimensions of cognitive load. To better assess the cognitive demands using the two-dimensional codes, we aggregated the codes by multiplying the two dimensions. For example, if an item was coded as having a cognitive load of task 2 level and dimension level 3 with an aggregate load of 6. We also included the difficulty rank provided by National Center for Education Statistics (NCES) for reference.

\begin{longtable}{m{1em}m{2em}m{2em}m{7em}m{3em}m{4em}m{2em}m{4em}}

        \caption{Descriptive and Coding Results of the Selected NAEP Assessment Tasks}\\
    \hline
    \centering
        ID & Grade & Type & Subject & Difficulty & \multicolumn{3}{c}{Cognitive Load} \\ 
        \cmidrule(lr){6-8}
        ~ & ~ & ~ & ~ & ~ & Dimension & Task& Aggregated \\ \hline
        1 & 4 & SCR & Physical Science & Medium & 3& 2 & 6 \\ 
        2& 4 & MC & Physical Science &Hard & 2 & 2 & 4 \\ 
        3 & 4 & MC & Physical Science &Hard  &  3& 2 & 6 \\ 
        4 & 4 & SCR & Physical Science & Hard & 3 & 3 & 9 \\ 
        5 & 4 &  MC& Physical Science & Hard & 3 & 2 & 6 \\ 
        6 & 4 & SCR & Physical Science & Hard & 2 & 3 & 6 \\
        7 & 8 & MC& Physical Science & Easy & 2 & 2 & 4 \\
        8 & 8 & MC & Physical Science & Medium & 3 & 2 & 6 \\ 
        9 & 8& SR & Physical Science & Easy & 2 & 2 & 4 \\ 
        10 & 8 & SR & Physical Science & Medium & 3 & 3 & 9 \\ 
        11 & 8 & SR & Physical Science & Easy & 2 & 2 & 4 \\ 
        12 & 8 & MC & Physical Science & Medium & 3 & 2 & 6 \\ 
        13 & 12 & MC & Physical Science & Medium & 3 & 2 & 6 \\ 
        14 & 12 & ECR & Physical Science & Hard & 3 & 3 & 9 \\ 
        15 & 12 & SCR & Physical Science & Hard & 3 & 4 & 12 \\ 
        16 & 12 & SCR & Physical Science & Hard & 3 & 4 & 12 \\ 
        17 & 12 & MC & Physical Science & Hard & 3 & 3 & 9 \\ 
        18 & 4 & MC & Life Science & Easy & 2 & 1 & 2 \\ 
        19 & 4 & MC & Life Science & Hard & 3 & 2 & 6 \\ 
        20 & 4 & MC & Life Science & Medium & 2 & 3 & 6 \\ 
        21 & 4 & MC & Life Science & Hard & 3 & 2 & 6 \\
        22 & 4 & MC & Life Science & Medium & 2 & 2 & 4 \\ 
        23 & 4 & MC & Life Science & Easy & 3 & 2 & 6 \\ 
        24 & 8 & MC & Life Science & Medium & 2 & 3 & 6 \\ 
        25 & 8 & ~MC & Life Science & Medium & 3 & 2 & 6 \\ 
        26 & 8 & SCR & Life Science & Hard & 3 & 3 & 9 \\ 
        27 & 8 & MC & Life Science & Easy & 2 & 1 & 2 \\ 
        28 & 8 & MC & Life Science & Medium & 2 & 3 & 6 \\ 
        29 & 8 & MC & Life Science & Easy & 2 & 2 & 4 \\ 
        30 & 12 & MC & Life Science & Easy & 2 & 1 & 2 \\ 
        31 & 12 & SCR & Life Science & Hard & 3 & 2 & 6 \\ 
        32 & 12 & MC & Life Science & Hard & 3 & 2 & 6 \\ 
        33 & 12 & MC & Life Science & Medium & 2 & 3 & 6 \\ 
        34 & 12 & MC & Life Science & Hard & 3 & 2 & 6 \\ 
        35 & 12 & SCR & Life Science & Hard & 3 & 2 & 6 \\ 
        36 & 12 & MC & Life Science & Easy & 2 & 1 & 2 \\ 
        37 & 12 & MC & Life Science & Easy & 2 & 2 & 4 \\ 
        38 & 4 & MC & Earth and Space Sciences & Medium & 3& 3 & 9 \\ 
        39 & 4 & SCR & Earth and Space Sciences & Medium & 3 & 3 & 9 \\ 
        40 & 4 & ECR & Earth and Space Sciences & Hard & 3 & 3 & 9 \\ 
        41 & 8 & SR & Earth and Space Sciences & Easy & 2 & 2 & 4 \\ 
        42 & 8 & MC & Earth and Space Sciences & Medium & 2 & 3 & 6 \\ 
        43 & 8 & MC & Earth and Space Sciences & Easy & 2 & 2 & 4 \\ 
        44 & 8 & ECR & Earth and Space Sciences & Medium & 2 & 3 & 6 \\ 
        45 & 8 & MC & Earth and Space Sciences & Easy & 2 & 2 & 4 \\ 
        46 & 8 & SCR & Earth and Space Sciences & Hard & 3 & 3 & 9 \\ 
        47 & 12 & MC & Earth and Space Sciences & Medium & 2 & 3 & 6 \\ 
        48 & 12 & MC & Earth and Space Sciences & Hard & 3 & 3 & 9 \\ 
        49 & 12 & MC & Earth and Space Sciences & Medium  & 2 & 3 & 6 \\
        50 & 12 & SCR & Earth and Space Sciences & Hard & 3 & 3 & 9 \\ 
        51 & 12 & MC & Earth and Space Sciences & Medium & 2 & 3 & 6 \\ \hline

    \end{longtable}

\subsection{Statistical Analysis}
To answer our research questions, we first asked ChatGPT and GPT-4 to answer the questions individually and used the scoring rubric to assign scores to their responses. This provides the performance of the ChatGPT and GPT-4. However, because some items require responders to observe simulations or images, to which altered cognitive demand significantly, we thus removed these items from our item pool. Eventually, only 47 items were included in the final comparison between students, ChatGPT and GPT-4. 

Due to privacy protection reasons, NAEP student data was restricted, and individual student performance was not accessible to researchers, which created challenges for us in comparing ChatGPT and GPT-4 and student performance. The available data on student performance are the average student ability scores for students who answered each item correctly calculated using Item Response Theory by NCES, and the percentage of students (POS) that correctly responded to individual items. Given this study is a secondary analysis and we directly employ the information provided by NCES, assessing score validity is beyond the scope of this study.

To answer the research question (a), we compared ChatGPT and GPT-4s’ performance with the POS and estimated the placement of ChatGPT and GPT-4 in the student population according to their ability. Specifically, if ChatGPT and GPT-4 correctly solve the problem $n$, for which POS is $a_n$, we estimate that ChatGPT and GPT-4 are ranked as the medium among students who correctly answer the question, and the placement is, 

         \[a_n/2 + (1- a_n) = 1 - a_n/2 \]

If ChatGPT and GPT-4 incorrectly solve the problem n, for which POS is an, we estimate that ChatGPT and GPT-4 are ranked as the medium among students who incorrectly answer the question, and the placement is,

            \[(1- a_n)/2\]
         
For example, assuming for an item, 86\% of students answered it correctly. If ChatGPT answers this item correctly, we rank ChatGPT as 1-.86/2= 57\%, suggesting that ChatGPT’s performance is higher than 57\% of students. If ChatGPT fails this item, we rank ChatGPT as (1-.86)/2= 7\%, suggesting that ChatGPT outperforms 7\% of students.  

To answer the research question (b), we conducted a crosstabulation analysis between ChatGPT and GPT-4 scores, required average student ability scores (RASAS) and the cognitive load of items by grade level, respectively. RASAS is the average ability of students who correctly answered the individual question, which are reported by The Nation’s Report Card. NEAP Crosstabulation analysis is a statistical method used to analyze the relationship between two or more categorical variables. The output of a crosstabulation is a contingency table, which presents the multivariate frequency distribution of the categories. Each cell in the table represents a count of the number of times a particular set of categories occurs together. By looking at the distribution of counts and/or percentages across the cells of the table, we can gain insights into the potential relationship between the variables.

\section{Results}
\subsection{Can ChatGPT or GPT-4 outperform humans on NAEP science assessments?}

Figures 1-3 show the percentages of students in Grades 4, 8, and 12, who scored below ChatGPT or GPT4 for each item. For some items, ChatGPT or GPT4 requests for more information were not excluded. The medians of the percentages were 83\%, 70\%, and 81\% for ChatGPT at Grade 4, 8, and 12, respectively, and 74\%, 71\%, and 81\% for GPT4 at Grades 4, 8, and 12, respectively. The interquartile ranges of the percentages were 20\%, 28\%, and 23\% for ChatGPT at Grades 4, 8, and 12, respectively, and 30\%, 8\%, and 21\% for GPT4 at Grades 4, 8, and 12, respectively. The results suggest that, compared with the sample of students who answered each individual item, ChatGPT and GPT4 usually perform above the median (Note that for a few items, ChatGPT or GPT4 requested more information and did not give an answer, so their responses were treated as missing values).

It can also be observed that ChatGPT and GPT4 performed similarly on most items. In particular, ChatGPT and GPT4 produced the same answer to 90\%, 75\%, and 94\% of items in Grades 4, 8, and 12. For items for which either ChatGPT or GPT4 requested additional information, the other one that did not request additional information often produced unsatisfactory answers.

\begin{figure}[htp]
\centering
\includegraphics[width=1\textwidth]{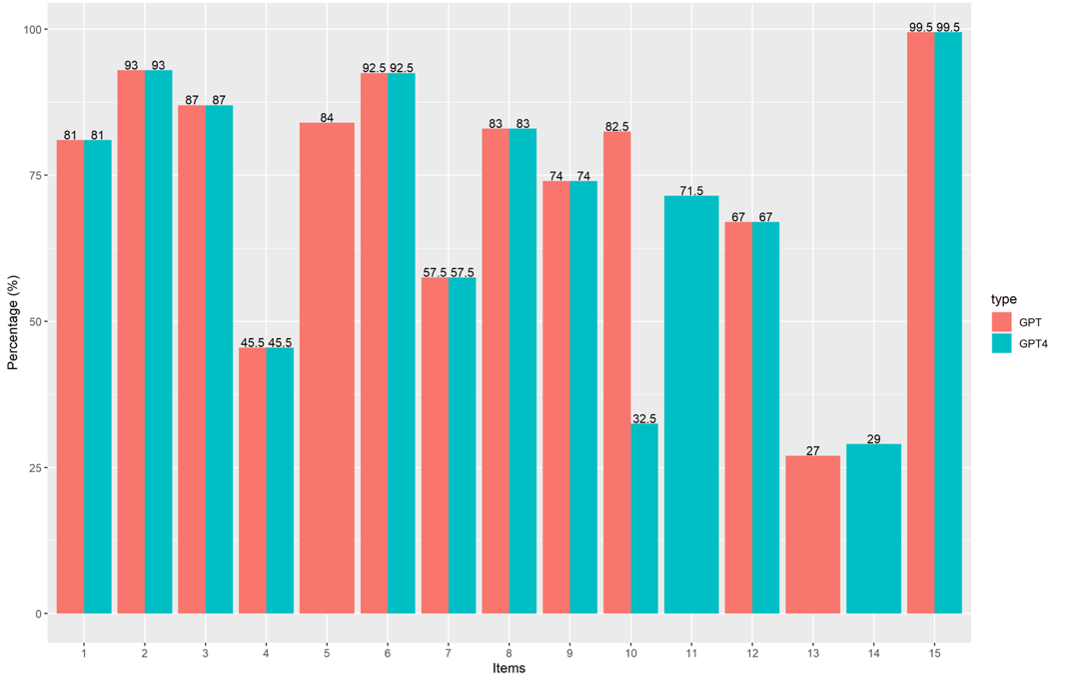}
\caption{percentage of students in Grade 4 scored below ChatGPT or GPT4 for each item}

\end{figure}

\begin{figure}[htp]
\centering
\includegraphics[width=1\textwidth]{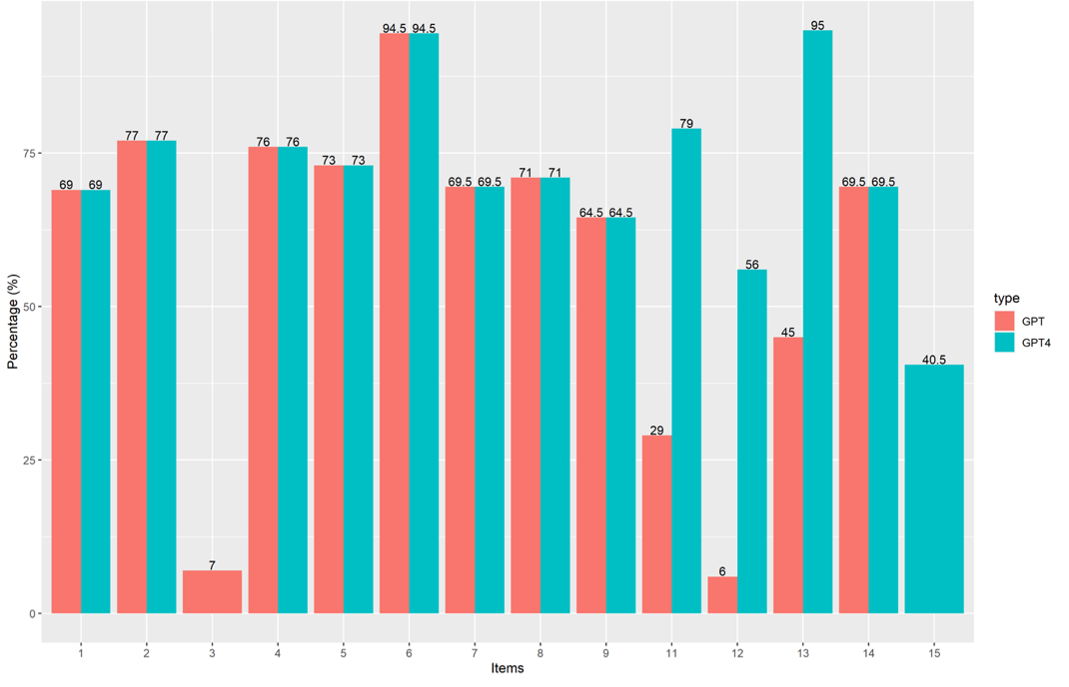}
\caption{ percentage of students in Grade 8 scored below ChatGPT or GPT4 for each item}

\end{figure}

\begin{figure}[htp]
\centering
\includegraphics[width=1\textwidth]{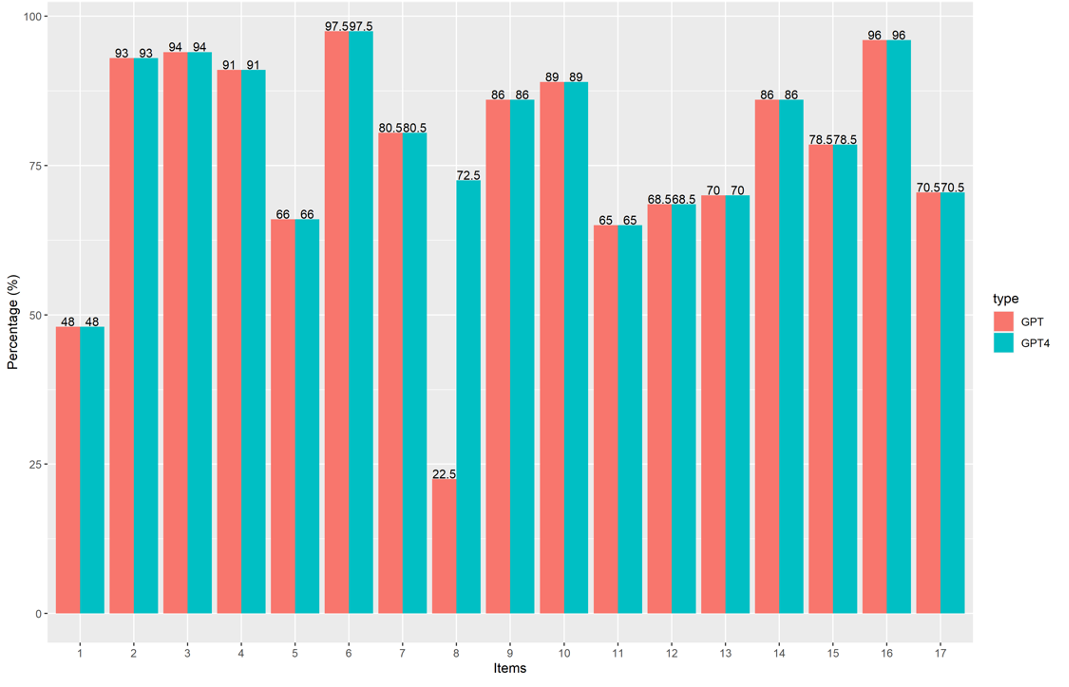}
\caption{ percentage of students in Grade 12 scored below ChatGPT or GPT4 for each item}

\end{figure}

\subsection{How ChatGPT and GPT-4 perform on NAEP science assessments by cognitive demand, }

\emph{compared to humans?}
The findings of Kendall’s $\tau_b$ correlation analysis were employed to elucidate the performance of ChatGPT and GPT-4 on NAEP science assessments compared to average students, segregated by cognitive demand levels.

In Grade 4, there was a significant positive correlation between the Required Average Student Ability Score (RASAS) and Cognitive Load (CL), $\tau_b$ (4) = .511, $p$ = .022, 95\% CI [.153, .750]. This implies that as cognitive demand increases, students who can correctly address the question need significantly increased average scores (or ability). Meanwhile, both ChatGPT (ChatGPTscore: $\tau_b$(4) = -.677, $p$ = .013, 95\% CI [-.850, -.373]) and GPT-4 (GPT4score: $\tau_b$ (4) = -.602, p = .026, 95\% CI [-.810, -.258]) demonstrated significant negative correlations with cognitive load, indicating that their performance declined with increased cognitive demand. In sum, the findings suggest that for the Grade 4 assessment tasks, 4th grade students, ChatGPT, and GPT-4 were all sensitive to cognitive load when solving the problems.

At Grade 8, the correlation between the RASAS and cognitive load (CL) was significant, $\tau_b$(8) = .600, $p$ = .005, 95\% CI [.294, .795], suggesting that the level of cognitive demand significantly impacts the average students’ performance by RASAS. For both GAI models, negative correlations were observed, albeit not statistically significant (ChatGPTscore: $\tau_b$(8) = 0, $p$ = 1.00, 95\% CI [-.407, .407]; GPT4score: $\tau_b$(8) = -.417, $p$ = .119, 95\% CI [-.704, -.012].The findings suggest that for Grade 8 assessment tasks, students’ performance was sensitive to cognitive load, while both ChatGPT and GPT-4 were not. The findings suggest that for the Grade 8 assessment tasks, students were sensitive but ChatGPT, and GPT-4 were insensitive to cognitive load when solving the problems.

In Grade 12, the RASAS and CL had a significant positive correlation, $\tau_b$(12) = .513, $p$ = .008, 95\% CI [.204, .729], indicating an increasing RASAS for students to successfully perform on tasks with increasing cognitive demand. Meanwhile, both GAI models again demonstrated negative correlations (ChatGPTscore: $\tau_b$(12) = -.162, $p$ = .477, 95\% CI [-.480, .194]; GPT4score: $\tau_b$(12) = -.246, $p$ = .279, 95\% CI [-.545, .107]), although these were not significant. The findings suggest that for Grade 12 assessment tasks, students’ performance was sensitive to cognitive load, while both ChatGPT and GPT-4 were not.

In summary, these findings suggest that for each of the three grade levels, higher average student ability scores are required on NAEP science assessments with increased cognitive demand. However, the performance of both ChatGPT and GPT-4 might not significantly impact the same conditions, except for the lowest grade 4.

    
\begin{table}[!ht]
    \centering
    \caption{Problem-Solving Abilities of Student, ChatGPT, and GPT-4 by Cognitive Demand}
  
    \begin{tabular}{m{5em}m{8em}m{6em}m{5em}m{5em}m{3em}}
    
    \hline
        Grade Level & Variables & Kendall’s $\tau_b$  & Significance (2-tailed) & \multicolumn{2}{c}{95\% Confidence Intervals $(2-tailed)^a$} \\ 
        \cmidrule(lr){5-6}
        ~ & ~ & ~ & ~ & Lower  & Upper \\ \hline
        4& RASAS-CL& $.511^*$ & \textbf{0.022} & 0.153 & 0.750 \\ 
        ~ & ChatGPTscore -CL & $-.677^*$ & \textbf{0.013} & -0.850 & -0.373 \\
        ~ & GPT4sore-CL& $-.602^*$ & \textbf{0.026}& -0.810 & -0.258 \\ 
        8 &RASAS-CL & .600** & \textbf{0.005} & 0.294 & 0.795 \\ 
        ~ & ChatGPTscore-CL & 0 & 1 & -0.407 & 0.407 \\ 
        ~ & GPT4sore-CL & -0.417 & 0.119 & -0.704 & -0.012 \\ 
        12 & RASAS-CL & .513** & \textbf{0.008} & 0.204 & 0.729 \\ 
        ~ & ChatGPTscore -CL & -0.162 & 0.477 & -0.480 & 0.194 \\ 
        ~ & GPT4sore-CL & -0.246 & 0.279 & -0.545 & 0.107 \\ \hline
    \end{tabular}
  
\end{table}
\footnotesize{a.Estimation is based on Fisher’s r-to-z transformation. RASAS = Required Average Student Ability Score; CL = Cognitive Load.}

\section{Conclusions and Discussion}
This study found that both ChatGPT and GPT-4 consistently outperformed the majority of students who answered each individual item in the NAEP science assessments. Additionally, ChatGPT and GPT-4 demonstrated a high degree of consistency in their performance. They produced the same answer for a significant proportion of the items across Grades 4, 8, and 12, with percentages of agreement reaching 90\%, 75\%, and 94\%, respectively. This similarity indicates that both models have a comparable level of accuracy and understanding when it comes to answering science-related questions. The research findings indicate that as the cognitive demand of NAEP science assessments increases, higher average student ability scores are required to correctly address the questions. This pattern was observed for Grade 4, Grade 8, and Grade 12 students respectively. However, neither ChatGPT nor GPT-4 demonstrated a significant correlation with cognitive load except for grade 4, indicating that their performance is not sensitive to cognitive demand. We suspect that the inconsistent performance of ChatGPT and GPT between grade 4 and grades 8 and 12 may be due to the small number of items. This finding suggests that for higher-grade level problem-solving, ChatGPT and GPT-4 may overcome the challenges brought by the increasing cognitive demands. The findings of this study have significant implications for educational practices and the role of GAI in assessments.

The consistent outperformance of both ChatGPT and GPT-4 compared to most students in the NAEP science assessments underscores the potential of GAI models to enhance educational outcomes and adds to the concern of misuse of LLMs in educational settings. These models demonstrate a high level of accuracy and understanding in addressing science-related questions, surpassing the performance levels of many students. The findings are consistent with prior research showing that ChatGPT and GPT-4 models possess high-order thinking and common-sense problem-solving ability (Bang et al., 2023; Bian et al., 2023). Given that NAEP represents the current education outcomes of the US across elementary to secondary schools, the findings are substantial and have amplified generalizability power, thus adding to the literature of GAI’s ability in scientific problem-solving. On the one hand, scholars are increasingly recognizing the capacity of GAI, particularly in the realm of artificial generative intelligence, as a transformative force in educational and learning contexts (Latif et al., 2023); this progress concurrently amplifies apprehensions regarding the potential misapplication of Large Language Models, including their use in outsourcing and related areas (Stokel-Walker, 2022).

The finding that ChatGPT and GPT-4 outperform students in solving science problems but are not sensitive to the cognitive demand required by these problems can be attributed to several potential reasons, with corresponding implications for education. One possible reason is that GAI models like ChatGPT and GPT-4 excel in pattern recognition and information retrieval (OpenAI, 2022). They are trained on extensive datasets and have access to vast amounts of information, allowing them to generate accurate responses. Their lack of sensitivity to cognitive demand demonstrates’ GAI’s potential to overcome the working memory that humans suffer when using higher-order thinking required by the problems.

The results of this study carry significant implications for the evolution of assessment practices within educational paradigms. Initially, there is an imperative for educators to overhaul traditional assessment practices (Zhai \& Wiebe, 2023). This reform is essential to ensure that these evaluations more accurately align with the evolving demands of skills and knowledge essential for student success in future societal contexts. The omnipresence of GAI in students’ lives indicates a shift away from emphasizing rote factual knowledge and basic skills, which are less pertinent in an era where such information is readily accessible through GAI technologies (Latif et al., 2023). Subsequently, given the noted insensitivity of GAI to cognitive load and its potential role as a tool in students’ future professional endeavors, it becomes crucial to recalibrate educational assessments. The focus of these assessments should pivot away from solely measuring cognitive intensity to a greater emphasis on creativity and the application of knowledge in novel contexts. This shift recognizes the growing importance of innovative thinking and problem-solving skills in a landscape increasingly influenced by advanced GAI technologies.

Consequently, education objectives should shift to nurturing critical thinking, problem-solving, creativity, and other cognitive abilities that GAI models may struggle to replicate. The finding underscores the need to strike a balance between utilizing GAI models and fostering human cognitive abilities. While GAI models can provide valuable assistance, they should not replace the development of higher-order cognitive abilities, encouraging students to think critically, analyze information, and apply knowledge to novel situations. For example, by focusing on inquiry-based instruction, students can actively explore scientific problems, develop hypotheses, conduct experiments, and analyze results. This process cultivates cognitive skills and ensures that students are not solely reliant on GAI models for problem-solving; instead, GAI can be an assistant to investigate science problems (Herdiska \& Zhai, 2023).

Education stakeholders should consider students’ development of metacognitive skills and ethical considerations related to GAI technology. Students should be equipped with the ability to critically evaluate and interpret the outputs of GAI models, understand their limitations and biases, and make informed decisions when using GAI tools. The capacity of ChatGPT and GPT-4 in this study empowers education to consider how to guide students in appropriate and ethical uses of GAI. Ethical awareness, including issues such as data privacy, algorithmic bias, and the impact of GAI on society, should also be integrated into education objectives. 

The capacity of GAI demonstrated in this study also indicates a significant shift, presenting challenges for which many educators are currently unprepared. This lack of preparedness among teaching professionals is a critical issue as they confront the task of integrating GAI into educational settings and guiding students in appropriate use of GAI, such as ChatGPT in education. To effectively navigate this transition, it is imperative for the educational system to offer comprehensive professional development opportunities. Such initiatives should aim to equip teachers with the necessary skills and knowledge, enabling them to guide students in harnessing the capabilities of GAI effectively and ethically. This approach is essential for educators to not only adapt to but also optimally exploit the transformative potential of GAI in educational contexts.

\section*{Limitations }
Despite the success of ChatGPT and GPT-4 on NAEP tasks, it is crucial to recognize the limitations of these models, such as their dependence on provided information and the need for additional context. Further research is needed to explore the full potential of GAI in education and ensure its ethical and effective implementation for the benefit of students and educators alike. Furthermore, it is important to note that when either ChatGPT or GPT-4 requested additional information for certain items, the model that did not request additional information often produced unsatisfactory answers. This finding suggests that these models heavily rely on the information provided to generate accurate responses. Consequently, the lack of additional context can significantly impact the quality of their answers, which aligns with prior research (Lee \& Zhai, 2023).

In addition, due to privacy protection reasons, NAEP student data was restricted, and individual student performance was not accessible to researchers, which created challenges for us in comparing ChatGPT and GPT-4 and student performance. The available data on student performance are the average student ability scores for students who answer each item correctly calculated using Item Response Theory, and the percentage of students that correctly responded to individual items. Future research should compare the individual students’ performance with large language models to better understand their difference and potentials. Further research is also needed to examine the inconsistent performance of ChatGPT and GPT between grade 4 and grades 8 and 12.

\section*{Acknowledgement}
The study was partially funded by National Science Foundation(NSF) (Award \# 2101104, PI: Zhai). Any opinions, findings, conclusions, or recommendations expressed in this material are those of the author(s) and do not necessarily reflect the views of the NSF.


\bibliographystyle{unsrtnat}
\bibliography{references}

\begin{hangparas}{.5in}{1}
Adiguzel, T., Kaya, M. H., \& Cansu, F. K. (2023). Revolutionizing education with AI: Exploring the transformative potential of ChatGPT. Contemporary Educational Technology, 15(3), ep429. 

Aktay, S., Seçkin, G., \& Uzunoğlu, D. (2023). ChatGPT in Education. Türk Akademik Yayınlar Dergisi (TAY Journal), 7(2), 378-406. 

Assaraf, N. (2022). Chatgpt: Optimizing language models for dialogue. 
\url{https://blog.cloudhq.net/openais-chatgpt-optimizing-language-models-for-dialogue/}

Bang, Y., Cahyawijaya, S., Lee, N., Dai, W., Su, D., Wilie, B., Lovenia, H., Ji, Z., Yu, T., \& Chung, W. (2023). A multitask, multilingual, multimodal evaluation of chatgpt on reasoning, hallucination, and interactivity. arXiv preprint arXiv:2302.04023.

D. A., \& Atwood, C. H. (2013). Implementation and assessment of Cognitive Load Theory (CLT) based questions in an electronic homework and testing system [10.1039/C3RP20153H]. Chemistry Education Research and Practice, 14(3), 247-256.\url{https://doi.org/10.1039/C3RP20153H}

Bergen, K. J., Johnson, P. A., de Hoop, M. V., \& Beroza, G. C. (2019). Machine learning for data-driven discovery in solid Earth geoscience. Science, 363(6433), eaau0323. 

Bergner, Y., \& von Davier, A. A. (2018, 2019/12/01). Process Data in NAEP: Past, Present, and Future. Journal of Educational and Behavioral Statistics, 44(6), 706-732. \url{https://doi.org/10.3102/1076998618784700}

Bian, N., Han, X., Sun, L., Lin, H., Lu, Y., \& He, B. (2023). ChatGPT is a Knowledgeable but Inexperienced Solver: An Investigation of Commonsense Problem in Large Language Models. arXiv preprint arXiv:2303.16421.

Brüggemann, T., Ludewig, U., Lorenz, R., \& McElvany, N. (2023). Effects of mode and medium in reading comprehension tests on cognitive load. Computers \& Education, 192, 104649.

Cao, C., Ding, Z., Lee, G.-G., Jiao, J., Lin, J., \& Zhai, X. (2023). Elucidating STEM Concepts through Generative AI: A Multi-modal Exploration of Analogical Reasoning. arXiv preprint arXiv:2308.10454. 

Daher, W., Diab, H., \& Rayan, A. (2023). Artificial Intelligence Generative Tools and Conceptual Knowledge in Problem Solving in Chemistry. Information, 14(7), 409. 

Estrella, S., Zakaryan, D., Olfos, R., \& Espinoza, G. (2020). How teachers learn to maintain the cognitive demand of tasks through Lesson Study. Journal of Mathematics Teacher Education, 23, 293-310. 

Feldon, D. F., Callan, G., Juth, S., \& Jeong, S. (2019, 2019/06/01). Cognitive Load as Motivational Cost. Educational Psychology Review, 31(2), 319-337. 
\url{https://doi.org/10.1007/s10648-019-09464-6}

Gerjets, P., Scheiter, K., \& Cierniak, G. (2009). The Scientific Value of Cognitive Load Theory: A Research Agenda Based on the Structuralist View of Theories. Educational Psychology Review, 21(1), 43-54. \url{https://doi.org/10.1007/s10648-008-9096-1}

Gupta, U., \& Zheng, R. Z. (2020). Cognitive Load in Solving Mathematics Problems: Validating the Role of Motivation and the Interaction among Prior Knowledge, Worked Examples, and Task Difficulty. European Journal of STEM Education, 5(1), 5. 

Hadie, S. N., \& Yusoff, M. S. (2016). Assessing the validity of the cognitive load scale in a problem-based learning setting. Journal of Taibah University Medical Sciences, 11(3), 194-202.

Herdiska, A., \& Zhai, X. (2023). Artificial Intelligence-Based Scientific Inquiry. In X. Zhai \& J. Krajcik (Eds.), Uses of Artificial Intelligence in STEM Education (pp. xxx-xxx). Oxford University Press. 

Ignjatović, A., \& Stevanović, L. (2023). Efficacy and limitations of ChatGPT as a biostatistical problem-solving tool in medical education in Serbia: a descriptive study. Journal of Educational Evaluation for Health Professions, 20, 28. \url{https://doi.org/10.3352/jeehp.2023.20.28}

Jordan, M. I., \& Mitchell, T. M. (2015). Machine learning: Trends, perspectives, and prospects. Science, 349(6245), 255-260. \url{https://science.sciencemag.org/content/349/6245/255.long}

Kung, T. H., Cheatham, M., Medenilla, A., Sillos, C., De Leon, L., Elepaño, C., Madriaga, M., Aggabao, R., Diaz-Candido, G., \& Maningo, J. (2023). Performance of ChatGPT on USMLE: Potential for AI-assisted medical education using large language models. PLOS Digital Health, 2(2), e0000198. 

Lagalante, M. C. R. (2023). High School Science Students’ Cognitive Load Using Virtual Reality Compared to Traditional Instruction Walden University]. 

Latif, E., Mai, G., Nyaaba, M., Wu, X., Liu, N., Lu, G., Li, S., Liu, T., \& Zhai, X. (2023, April 01, 2023). AGI: Artificial General Intelligence for Education. arXiv:2304.12479. \url{https://doi.org/10.48550/arXiv.2304.12479}

Lee, G.-G., \& Zhai, X. (2023). NERIF: GPT-4V for Automatic Scoring of Drawn Models. arXiv preprint arXiv:2311.12990. 

Li, P.-H., Lee, H.-Y., Cheng, Y.-P., Starčič, A. I., \& Huang, Y.-M. (2023). Solving the Self-regulated Learning Problem: Exploring the Performance of ChatGPT in Mathematics. In Y.-M. Huang \& T. Rocha, Innovative Technologies and Learning Cham.

Lim, H., \& Sireci, S. G. (2017). Linking TIMSS and NAEP assessments to evaluate international trends in achievement. Education Policy Analysis Archives, 25, 11. \url{https://doi.org/10.14507/epaa.25.2682}

McCormick, M. (2016). Exploring the Cognitive Demand and Features of Problem Solving Tasks in Primary Mathematics Classrooms. Mathematics Education Research Group of Australasia. 

Najafabadi, M. M., Villanustre, F., Khoshgoftaar, T. M., Seliya, N., Wald, R., \& Muharemagic, E. (2015). Deep learning applications and challenges in big data analytics. Journal of big data, 2(1), 1-21. 

National Assessment Governing Board. (2019). Science framework for the 2019 national assessment of educational progress. 

NGSS Lead States. (2013). Next generation science standards: For states, by states. National Academies Press. 

OpenAI. (2022). ChatGPT: Optimizing Language Models for Dialogue.\url{ https://openai.com/blog/chatgpt/}

OpenAI.(2023).GPT4.\url{https://openai.com/research/gpt-4}

Orrù, G., Piarulli, A., Conversano, C., \& Gemignani, A. (2023a). Human-like problem-solving abilities in large language models using ChatGPT. Frontiers in Artificial Intelligence, 6, 1199350. \url{https://www.ncbi.nlm.nih.gov/pmc/articles/PMC10244637/pdf/frai-06-1199350.pdf}

Orrù, G., Piarulli, A., Conversano, C., \& Gemignani, A. (2023b). Human-like problem-solving abilities in large language models using ChatGPT. Frontiers in artificial intelligence, 6. \url{https://doi.org/10.3389/frai.2023.1199350}

Paas, F., Renkl, A., \& Sweller, J. (2003). Cognitive load theory and instructional design: Recent developments. Educational Psychologist, 38(1), 1-4. 

Paas, F., \& Van Merriënboer, J. J. G. (2020). Cognitive-Load Theory: Methods to Manage Working Memory Load in the Learning of Complex Tasks. Current Directions in Psychological Science, 29(4), 394-398. \url{https://doi.org/10.1177/0963721420922183}

Pengelley, J., Whipp, P. R., \& Rovis-Hermann, N. (2023, 2023/06/20). A Testing Load: Investigating Test Mode Effects on Test Score, Cognitive Load and Scratch Paper Use with Secondary School Students. Educational Psychology Review, 35(3), 67. \url{https://doi.org/10.1007/s10648-023-09781-x}

Prisacari, A. A., \& Danielson, J. (2017). Computer-based versus paper-based testing: Investigating testing mode with cognitive load and scratch paper use. Computers in Human Behavior, 77, 1-10. 

Rosenfeld, S. (2011). Common sense: A political history. Harvard University Press. 

Seetharaman, R. (2023). Revolutionizing Medical Education: Can ChatGPT Boost Subjective Learning and Expression? Journal of Medical Systems, 47(1). \url{https://doi.org/10.1007/s10916-023-01957-w}

Sinha, R. K., Roy, A. D., Kumar, N., Mondal, H., \& Sinha, R. (2023). Applicability of ChatGPT in assisting to solve higher order problems in pathology. Cureus, 15(2). 

Stokel-Walker, C. (2022). AI bot ChatGPT writes smart essays-should academics worry? Nature. 

Sweller, J. (2011). Cognitive load theory. In Psychology of learning and motivation (Vol. 55, pp. 37-76). Elsevier. 

Tekkumru‐Kisa, M., Stein, M. K., \& Schunn, C. (2015). A framework for analyzing cognitive demand and content‐practices integration: Task analysis guide in science. Journal of Research in Science Teaching, 52(5), 659-685. 

The Nation’s Report Card. (2022). Question Tool. Retrieved May 13 from \url{https://www.nationsreportcard.gov/nqt/searchquestions}

Tugtekin, U., \& Odabasi, H. F. (2022). Do Interactive Learning Environments Have an Effect on Learning Outcomes, Cognitive Load and Metacognitive Judgments? Education and Information Technologies, 27(5), 7019-7058. https://doi.org/10.1007/s10639-022-10912-0 

Wang, T., Li, M., Thummaphan, P., \& Ruiz-Primo, M. A. (2017, 2017/10/02). The Effect of Sequential Cues of Item Contexts in Science Assessment. International Journal of Testing, 17(4), 322-350. \url{https://doi.org/10.1080/15305058.2017.1297818}

Williams, A. E. (2023). Has OpenAI Achieved Artificial General Intelligence in ChatGPT? Artificial Intelligence and Applications. \url{https://doi.org/10.47852/bonviewaia3202751}

Zeng, F. (2023). Evaluating the Problem Solving Abilities of ChatGPT.

Zhai, X. (2022). ChatGPT user experience: Implications for education. Available at SSRN 4312418. 

Zhai, X. (2023). ChatGPT and AI: The Game Changer for Education. SSRN. \url{https://doi.org/https://ssrn.com/abstract=4389098}

Zhai, X., \& Pellegrino, J. (2023). Large-Scale Assessment in Science Education. In N. G. Lederman, D. L. Zeidler, \& J. S. Lederman (Eds.), Handbook of research on science education (Vol. III, pp. 1045-1098). Foutledge.

Zhai, X., \& Wiebe, E. (2023). Technology-Based Innovative Assessment. In C. J. Harris, E. Wiebe, S. Grover, \& J. W. Pellegrino (Eds.), Classroom-Based STEM Assessment (pp. 99-125). Community for Advancing Discovery Research in Education, Education Development Center, Inc. 

Zhai, X., Yin, Y., Pellegrino, J. W., Haudek, K. C., \& Shi, L. (2020). Applying machine learning in science assessment: a systematic review. Studies in Science Education, 56(1), 111-151.







\end{hangparas}
\end{document}